%% file: main.tex
\pdfoutput=1

\documentclass{article}
\usepackage{spconf,amsmath,graphicx}
\usepackage{amsfonts}
\usepackage{tikz}
\usetikzlibrary{calc}
\usepackage{epstopdf}
\usepackage{graphicx}
\newcommand{\caja}[4][1]{{%
    \renewcommand{\arraystretch}{#1}%
    \begin{tabular}[#2]{@{}#3@{}}%
      #4%
    \end{tabular}%
    }}

\usepackage{color}

\graphicspath{{grf/}}

\title{SIGNER-INDEPENDENT FINGERSPELLING RECOGNITION \\ WITH
DEEP NEURAL NETWORK ADAPTATION}
\name{Taehwan Kim, Weiran Wang, Hao Tang, Karen Livescu\sthanks{This research was supported by NSF grant NSF-1433485.  The opinions expressed in this work are those of the authors and do not necessarily reflect the views of the funding agency.}}
\address{Toyota Technological Institute at Chicago, USA \\
         {\small \tt \{taehwan,weiranwang,haotang,klivescu\}@ttic.edu}}
\begin{document}
%
\maketitle
\begin{abstract}
We study the problem of recognition of fingerspelled letter sequences in American Sign Language in a signer-independent setting.  Fingerspelled sequences are both challenging and important to recognize, as they are used for many content words such as proper nouns and technical terms. Previous work has shown that it is possible to achieve almost $90$\% accuracies on fingerspelling recognition in a signer-dependent setting.  However, the more realistic signer-independent setting presents challenges due to significant variations among signers, coupled with the dearth of available training data.  We investigate this problem with approaches inspired by automatic speech recognition.  
We start with the best-performing approaches from prior work, based on tandem models and segmental conditional random fields (SCRFs), with features based on deep neural network (DNN) classifiers of letters and phonological features.  Using DNN adaptation, we find that it is possible to bridge a large part of the gap between signer-dependent and signer-independent performance.   Using only about 115 transcribed words for adaptation from the target signer, we obtain letter accuracies of up to $82.7$\% with frame-level adaptation labels and $69.7$\% with only word labels.
\end{abstract}

\input{intro}



\input{method}

\input{experiments}

\input{conclusions}


\clearpage
\begin{small}
\bibliographystyle{IEEEbib}
\bibliography{icassp16b}
\end{small}
\end{document}

%% file: intro.tex
\section{INTRODUCTION}
\label{sec:intro}

Automatic sign language recognition is a nascent technology that has the potential to improve the ability of Deaf and hearing individuals to communicate, as well as Deaf individuals' ability to take full advantage of modern information technology.  For example, online sign language video blogs and news\footnote{E.g., \tt{http://ideafnews.com, http://aslized.org}.} are currently almost completely unindexed and unsearchable as they include little accompanying annotation.  

Research on this problem has included both speech-inspired approaches and computer vision-based techniques, using either/both video and depth sensor input~\cite{dreuw,zaki,bowden,liw,theo,vog,kim2012,kim2013,forster2013improving}.
We focus on recognition from video, for applicability to existing recordings.  Before the technology can be applied ``in the wild'', it must overcome challenges posed by visual nuisance parameters (e.g., lighting, occlusions) and signer variation.  Annotated data sets for this problem are scarce, in part due to the need to recruit signers and skilled annotators.  


We consider American Sign Language (ASL), in particular the fingerspelling component:  the spelling out of a word as a sequence of handshapes or hand trajectories corresponding to individual letters.  
Fig.~\ref{fig:ex} gives example fingerspelling sequences.  Fingerspelling accounts for roughly 12-35\% of ASL~\cite{Padden} and is typically used for proper nouns or borrowings from English, 
which can often be the most important content words.  
Some aspects of fingerspelling can be characterized through the phonology of handshape~\cite{bren,john}, which can be described in terms of phonological features.
Most prior research on fingerspelling recognition has focused on constrained tasks such as single-letter or handshape classification or word recognition from a known vocabulary~\cite{athitsos2004boostmap,Ricco-Tomasi-09,Tsechpenakis-et-al-06-coupling,Bowden-Sarhadi-02,Goh-Holden-06,Roussos:JMLR.13,PugBow2011}.  For the unconstrained letter sequence recognition problem, Kim et al.~\cite{kim2012,kim2013} obtained $\sim90$\% average letter accuracies in a signer-dependent setting, using either tandem hidden Markov models (HMMs) or segmental conditional random fields (SCRFs), with features from neural network classifiers of letters and phonological features.  That work used the largest video data set of which we are aware containing unconstrained, connected fingerspelling, consisting of four signers each signing 600 word tokens for a total of $\sim350$k image frames. 

In this paper we consider the problem of {\it signer-indepen-dence in unconstrained fingerspelling sequence recognition}, in the context of limited training data.  
Prior work has address signer adaptation for large-vocabulary German Sign Language recognition~\cite{forster2013improving}, but to our knowledge this paper is the first to address adaptation for fingerspelling.  We 
investigate approaches to signer-independence including speed normalization and neural network adaptation.  The adaptation techniques are largely borrowed from speech recognition research, but the application is quite different in that the overall amount of data is much smaller and the types of variation are different.  We find that the simple signer normalization is ineffective, while DNN adaptation is very effective.  

%% file: method.tex
\section{METHODS}
\label{sec:method}
\vspace{-.1in}

\input{examplefig}

The task is to convert a video (a sequence of images), as in Fig.~\ref{fig:ex}, to a sequence of letters.  The segmentation into letters is unknown, so this is a sequence prediction task analogous to connected phone or word recognition.  We start with the recognition approaches that have achieved the best prior results on this task~\cite{kim2012,kim2013}, with updates for improved performance with deeper neural networks.  We next briefly describe the recognizers, the neural network classifiers and adaptation.

\vspace{-.2in}
\subsection{Recognizers}
\vspace{-.05in}
The first recognizer is a tandem model~\cite{ellis} based on~\cite{kim2012}.  Frame-level features are fed to neural network classifiers, one of which predicts the frame's letter label and six others which predict handshape phonological features.\footnote{See~\cite{bren,john,kim2012} for details of the phonological features.}  Classifier outputs are concatenated with the image features, after a dimensionality reduction, and input to a hidden Markov model (HMM) recognizer with Gaussian mixture observation densities.

The second recognizer is a segmental CRF (SCRF) model based on~\cite{kim2013}.  SCRFs~\cite{sarawagisemi,zweig} are conditional log-linear models with feature functions that can be based on variable-length segments of input frames, allowing for great flexibility in defining feature functions.  
As in~\cite{kim2013}, we use an SCRF to rescore lattices produced by a baseline frame-based recognizer (in this case, the tandem model).  We use the same feature functions as in~\cite{kim2013}, which include language model features, a feature that measures agreement with the baseline recognizer, means of letter/phonological feature neural network classifier outputs over each segment, and ``peak detection'' features that measure the dynamics of each segment.

Finally, we also use a first-pass decoding SCRF from Tang {\it et al.}~\cite{tang2015}, which is independent of any frame-based recognizer.  We use the same feature functions as in~\cite{tang2015}, namely average DNN outputs over each segment, samples of DNN outputs within the segment, duration and bias, all lexicalized.

\input{dnn}

%% file: examplefig.tex
\begin{figure*}[!th]
\begin{tikzpicture}

  \foreach \pos/\imnum in {5.92/1, 7.885/11}
  \node at ($(4*\pos,1.6)$) {\includegraphics[width=2cm]{figs/ex_andy\imnum}};
  \node at ($(22.18,1.6)$) {\includegraphics[width=2cm]{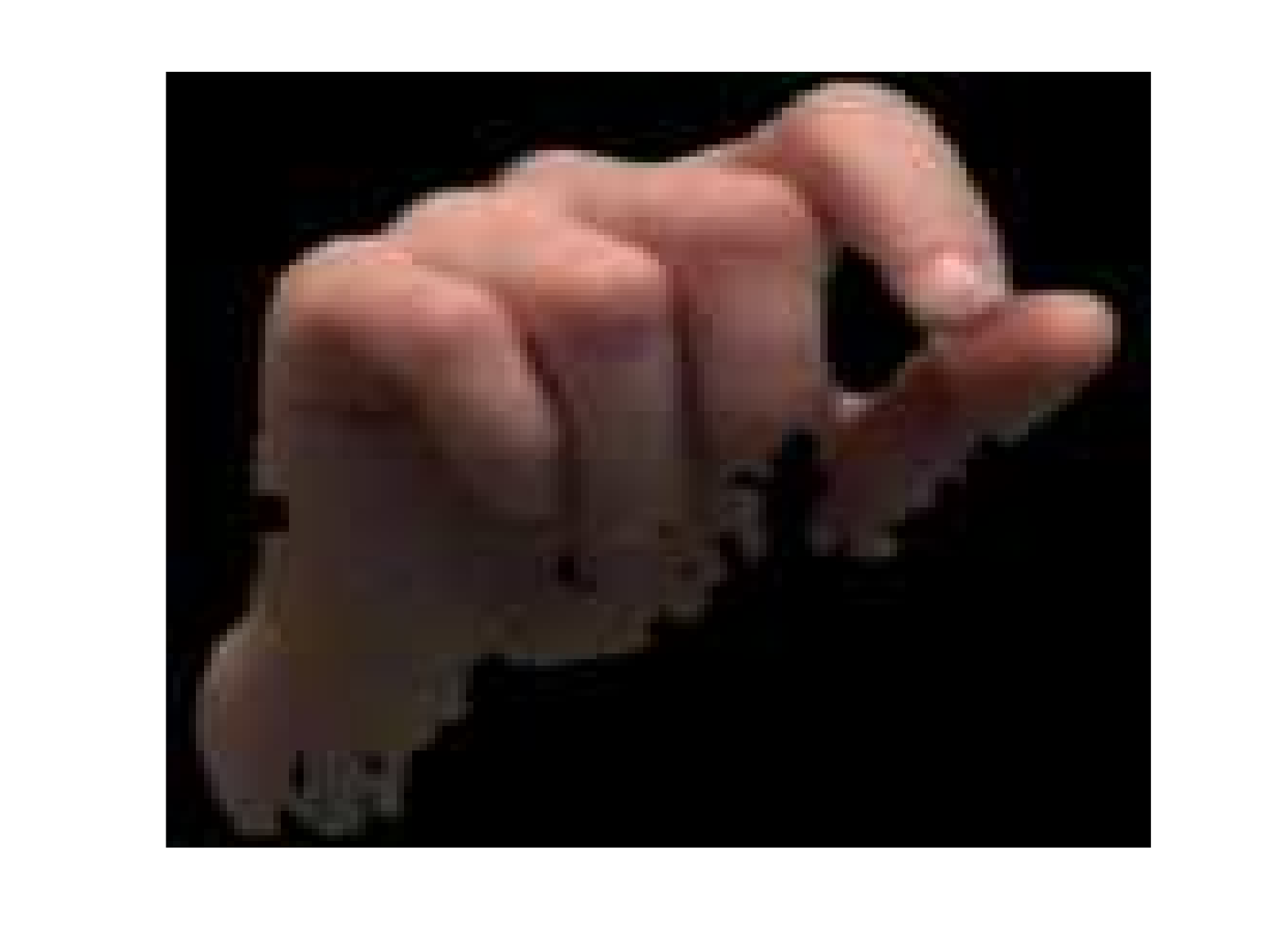}};
  \node at ($(20.68,1.6)$) {\includegraphics[width=2cm]{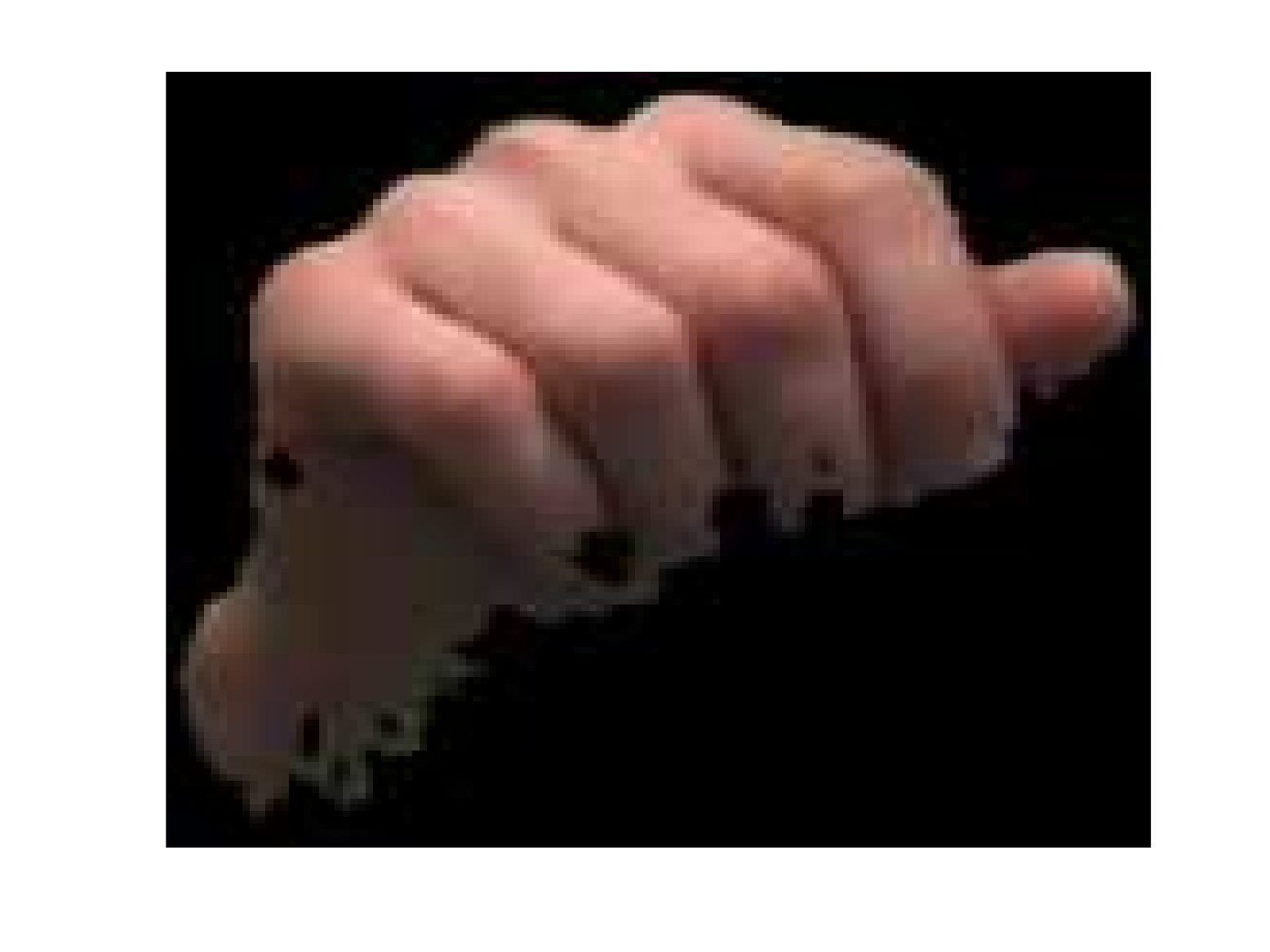}};
  
    \node at ($(32.84,1.6)$) {\includegraphics[width=2cm]{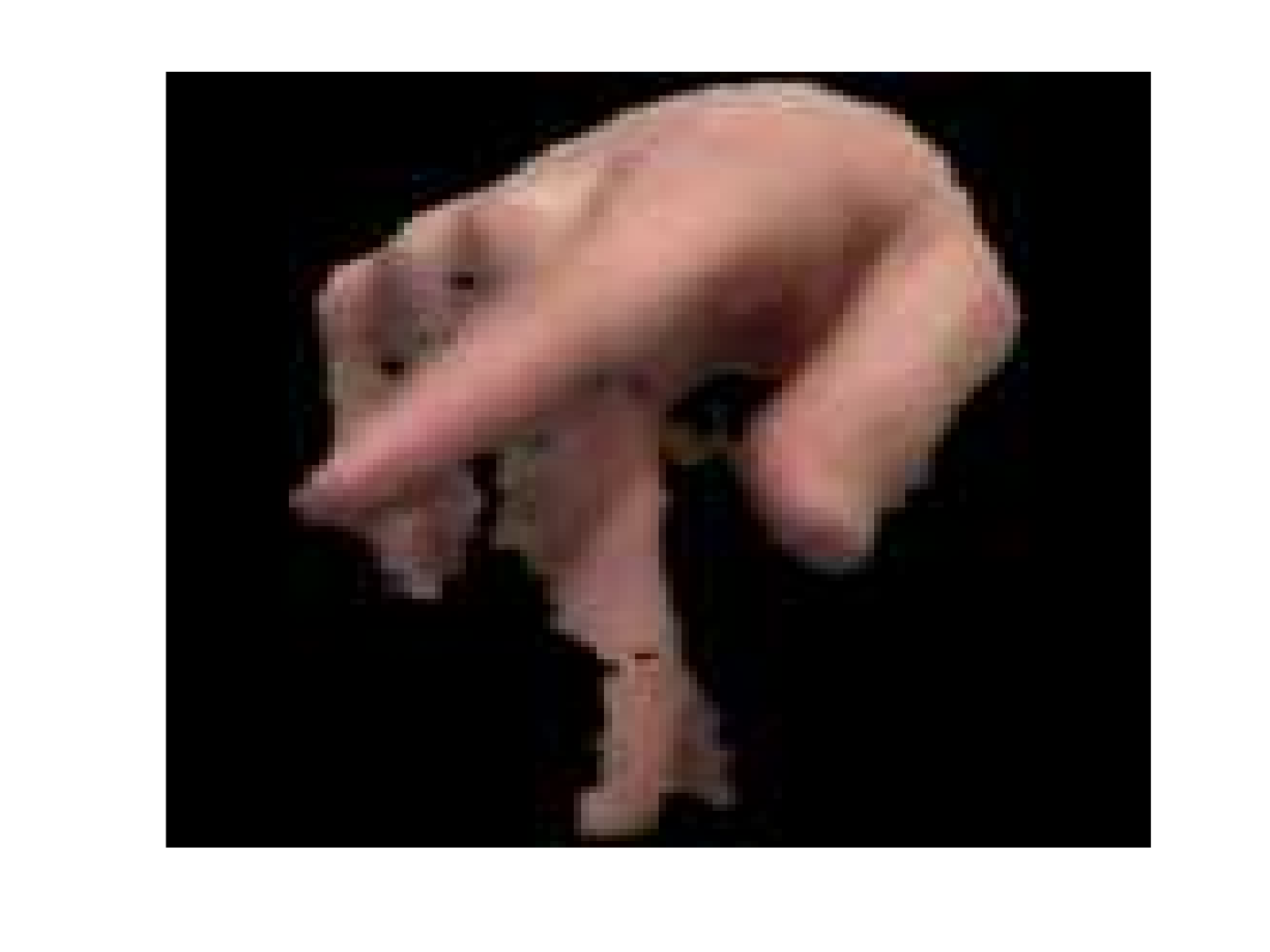}};
  \node at ($(34.14,1.6)$) {\includegraphics[width=2cm]{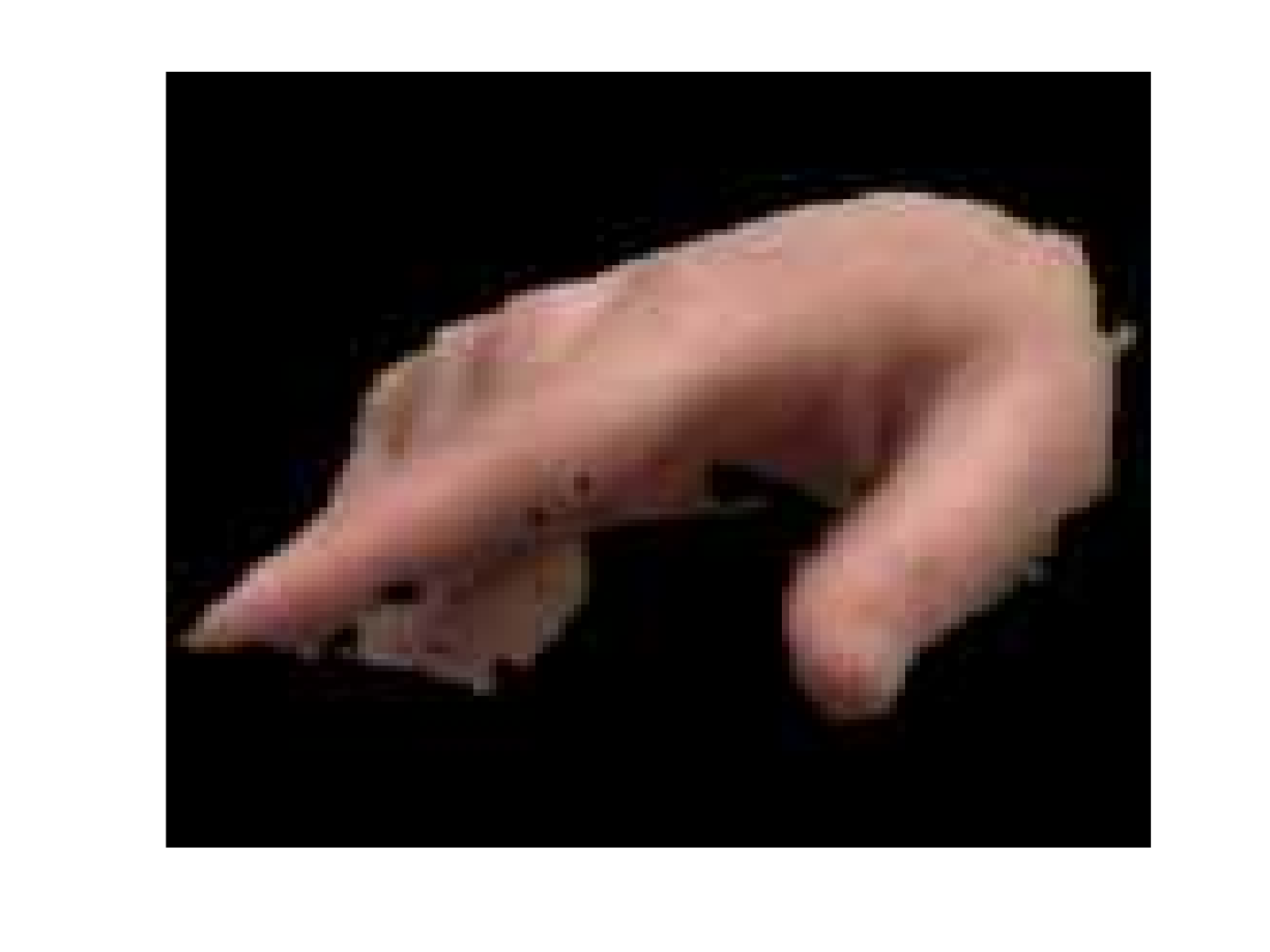}};
  
  
  
  
  \draw [black,thick] (20,0)--(24.30,0)--(24.30,.8)--(20,.8);
  \node at (22.20,.3) {$<$s$>$};
  \draw [black,thick] (34.96,0)--(30.94,0)--(30.94,.8)--(34.96,.8);
  \node at (32.94,.3) {$<$/s$>$};
  
    \foreach \sta/\sto/\letter/\imnum in { 6.4787/6.7356/U/4, 6.075/6.4787/T/2,   7.0659/7.3962/I/8, 6.7356/7.0659/L/6, 7.3962/7.735/P/10} {
    \draw [black,thick] ($(4*\sta,0)$) rectangle ($(4*\sto,.8)$);
    \node at ($(4*\sta,0.3)!.5!(4*\sto,0.3)$) {\letter};
    \node at ($(4*\sta,1.6)!.5!(4*\sto,1.6)$)
    {\includegraphics[width=2cm]{figs/ex_andy\imnum}};
    \node at ($(4*\sta,2.3)!.5!(4*\sto,2.3)$) {$\ast$};
  }
  
  
    \foreach \sta/\sto/\letter/\imnum in { 5.6075/6.2584/T/2, 6.2584/7.0629/U/4, 7.0629/7.7968/L/6, 7.7968/8.5541/I/8, 8.5541/9.2075/P/10} {
    \draw [black,thick] ($(4*\sta,2.5)$) rectangle ($(4*\sto,3.3)$);
    \node at ($(4*\sta,2.8)!.5!(4*\sto,2.8)$) {\letter};
    \node at ($(4*\sta,4.1)!.5!(4*\sto,4.1)$)
    {\includegraphics[width=2cm]{figs/ex_drucie\imnum}};
    \node at ($(4*\sta,4.8)!.5!(4*\sto,4.8)$) {$\ast$};
  }
    \foreach \pos/\imnum in {5.5575/1, 6.2818/3, 7.0455/5, 7.79/7, 8.5591/9, 9.2575/11}
  \node at ($(4*\pos,4.1)$) {\includegraphics[width=2cm]{figs/ex_drucie\imnum}};
  
    \draw [black,thick] (20,2.5)--(22.43,2.5)--(22.43,3.3)--(20,3.3);
  \node at (21.1,2.8) {$<$s$>$};
  \draw [black,thick] (37.83,2.5)--(36.83,2.5)--(36.83,3.3)--(37.83,3.3);
  \node at (37.48,2.8) {$<$/s$>$};
  
  \node at ($(20.7,4.1)$) {\includegraphics[width=2cm]{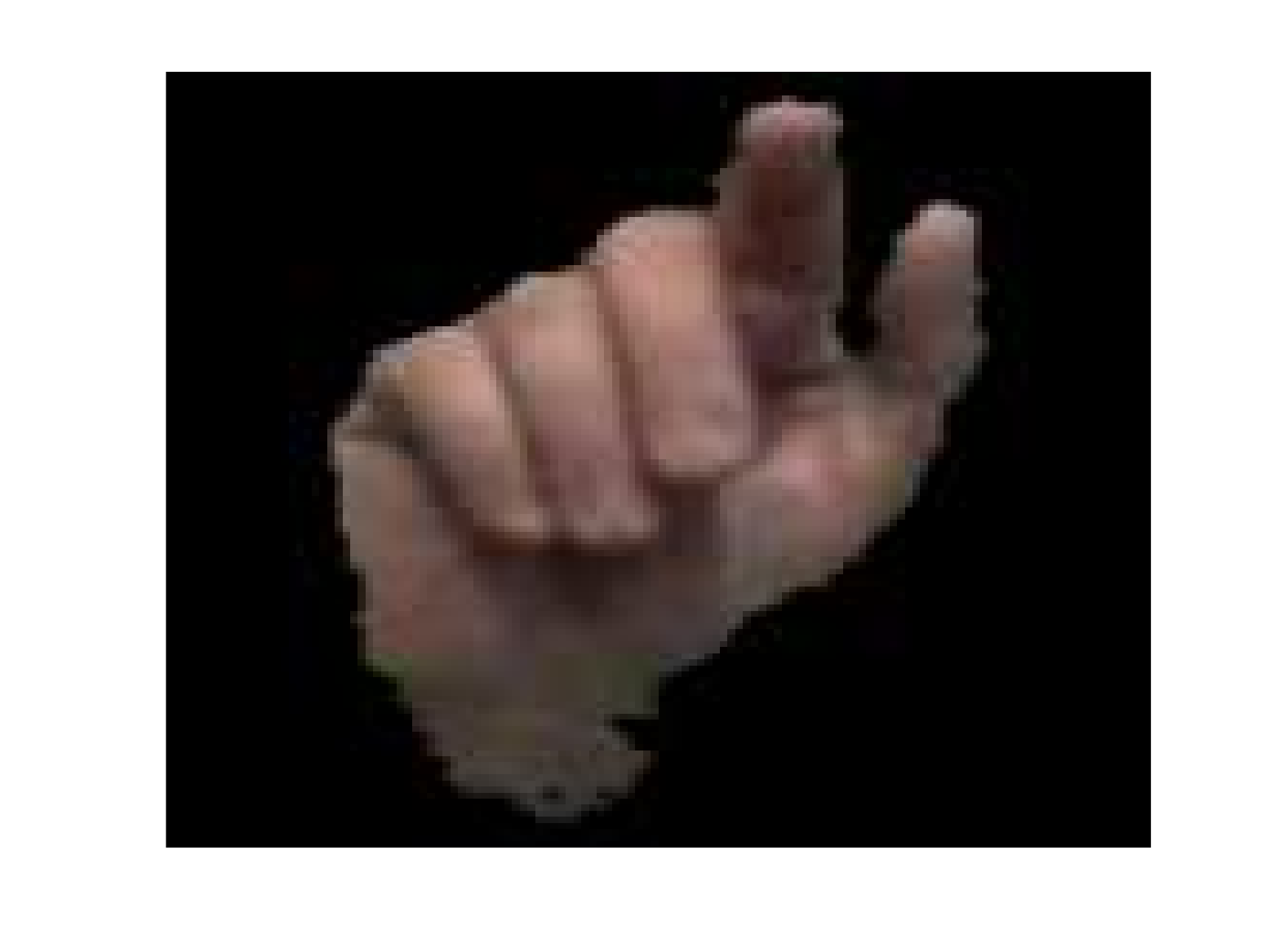}};

  
  

  \foreach \sta/\sto/\letter in { 5/5/0, 6/6/1, 7/7/2, 8/8/3, 9/9/4, 9.3/9.3/Time(sec)} 
    \node at ($(4*\sta,-.4)!.5!(4*\sto,-.4)$) {\letter};

\end{tikzpicture}
\vspace{-.2in}
\caption{Images and ground-truth segmentations of the fingerspelled word `TULIP' produced by two signers. Image frames are sub-sampled at the same rate from both signers to show the true relative speeds.  Asterisks indicate manually annotated peak frames for each letter.  ``$<$s$>$'' and ``$<$/s$>$'' denote non-signing intervals before/after signing.}
\label{fig:ex}
\end{figure*}

%% file: dnn.tex
\vspace{-.15in}
\subsection{DNN adaptation}
\label{sec:dnn}
\vspace{-.05in}

\begin{figure*}[th]
\centering
\includegraphics[width=.8\linewidth]{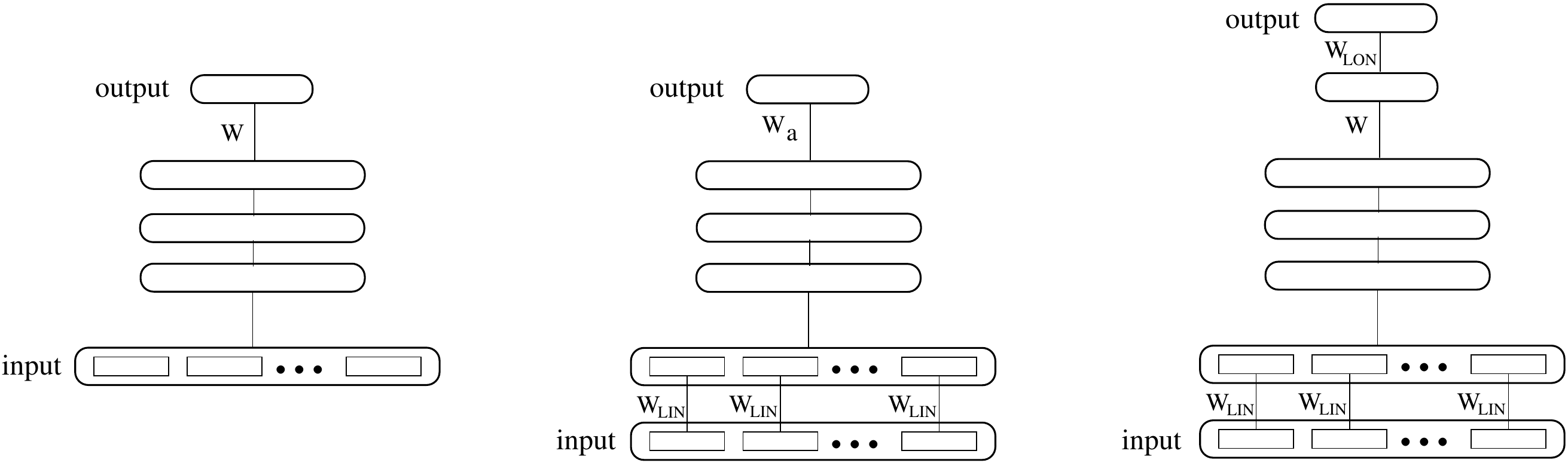}
\vspace{-.1in}
\caption{Left:  Unadapted DNN classifier; middle:  adaptation via linear input network and output layer updating (LIN+UP); right: adaptation via linear input network and linear output network (LIN+LON).}
\label{fig:dnn-adapt}
\end{figure*}

The DNNs 
are first trained in a signer-independent way on all but the test signer, using an L2-regularized cross-entropy loss.  The inputs are the image features concatenated over a multi-frame window,
which are fed through several fully connected layers followed by a softmax output layer.
Inspection of data such as Fig.~\ref{fig:ex} reveals the main sources of signer variation:  speed, hand appearance, and non-signing motion variation before/after signing.  The speed variation is large, with a factor of $1.8$ between the fastest and slowest signers.  In the absence of adaptation data, we consider a simple speed normalization:  We augment the training data with resampled image features, at 0.8x and 1.2x the original frame rate.

If we have access to some labeled data from the test signer, but not a sufficient amount for training full signer-specific DNNs, we can apply adaptation.  A number of DNN adaptation approaches have been developed (e.g.,~\cite{liao2013speaker,abdel2013fast,swietojanski2014learning,doddipatla2014speaker}).  We first consider two simple approaches based on linear input networks (LIN) and linear output networks (LON)~\cite{Neto_95a,Yao_12a,li2010}, shown in Fig.~\ref{fig:dnn-adapt}.  Most of the network parameters are fixed; only a limited set of weights at the input and output layers are learned. 
In the first approach (LIN+UP in Fig.~\ref{fig:dnn-adapt}), we apply a single affine transformation $W_{\text{LIN}}$ to the 
static features at each frame (before concatenation)
and feed the result to the trained signer-independent DNNs. We jointly learn $W_{\text{LIN}}$ and adapt the last (softmax) layer weights by minimizing the same cross-entropy loss
on the adaptation data, 
and ``warm-start'' the softmax layer with the learned signer-independent weights.
The second approach (LIN+LON in Fig.~\ref{fig:dnn-adapt}) uses the same input adaptation layer, but rather than adapting the softmax weights, it removes the softmax output activation and adds a new softmax output layer $W_{\text{LON}}$ for the test signer,
trained jointly with the same cross-entropy loss. Finally, we also consider adaptation by fine-tuning; that is, updating all of the DNN weights on adaptation data starting from the signer-independent weights.  The adaptation can use either ground-truth frame-level letter labels (using human annotation as described in~\cite{kim2012}) or labels obtained by forced alignment if only word labels are available.


%% file: experiments.tex
\vspace{-.2in}
\section{EXPERIMENTS}
\label{sec:experiments}
\vspace{-.1in}

\input{table}

We use the ASL video data set of~\cite{kim2013}, comprising four signers each fingerspelling 600 word tokens consisting of two repetitions of a 300-word list, including common English words, names, and foreign words. 
Annotators marked the peak of articulation of each letter, and the annotations were converted to a ``ground-truth'' frame labeling by assuming that the letter boundaries occur mid-way between peaks.  Following \cite{kim2013}, the hand portion of each image is extracted via hand detection and segmentation using a signer-specific Gaussian color model, followed by suppression of irrelevant pixels.  The extracted hand images are resized to 128 $\times$ 128 and Histogram of Gradient (HoG) \cite{hog} features are extracted using multiple spatial grids (4 $\times$ 4, 8 $\times$ 8, and 16 $\times$ 16), followed by dimensionality reduction with principal components analysis (PCA).  


\vspace{-.15in}
\subsection{Frame classification}
\vspace{-.05in}

The initial unadapted signer-independent DNNs are trained on all but the test signer for each of the seven tasks (letters and the six phonological features).  The input is the $128$-dimensional HoG features
concatenated over a $21$-frame window, and the networks have three hidden layers of $3000$ ReLUs~\cite{Zeiler_13a}.  Cross-entropy training is done with a 
weight decay penalty of $10^{-5}$ via stochastic gradient descent (SGD) over $100$-sample minibatches for up to $30$ epochs, with dropout~\cite{Srivas_14a} at a rate of $0.5$ at each hidden layer, fixed momentum of $0.95$, and initial learning rate of $0.01$, which is halved when held-out accuracy stops improving.  These hyperparameters were tuned on held-out (signer-independent) data in initial experiments, not reported here in the interest of space.  We pick the best-performing epoch on held-out data.  


\begin{figure}\centering
\centering
\vspace{-.06in}
\hspace{-.2in}\includegraphics[width=0.92\linewidth]{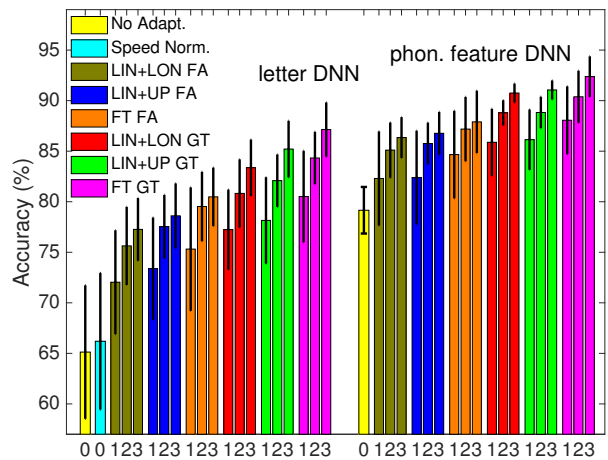}
\vspace{-.15in}
\caption{DNN frame accuracies with and without normalization/adaptation.  The horizontal axis labels indicate the amount of adaptation data (0, 1, 2, 3 = none, 5\%, 10\%, 20\% of the test signer's data, corresponding to no adaptation, $\sim29$, $\sim58$, and $\sim115$ words).  GT = ground truth labels; FA = forced alignment labels; FT = fine-tuning.}
\label{fig:adapt}
\end{figure}


We next consider DNN normalization and adaptation with different types and amounts of supervision.  For LIN+UP and LIN+LON, we adapt by running SGD over minibatches of $100$ samples with a fixed momentum of $0.9$ for up to $20$ epochs, with initial learning rate of $0.02$ (which is halved when accuracy stops improving on the adaptation data).  For fine-tuning, we use the same SGD procedure as for the signer-independent DNNs.  We pick the epoch with the highest accuracy on the adaptation data. 
The resulting frame accuracies are given in Fig.~\ref{fig:adapt}.  In addition, Fig.~\ref{fig:adapt} includes the result of speed normalization for the case of letter classification.  Speed normalization provides consistent but very small improvements, while adaptation gives large improvements in all settings.
LIN+UP slightly outperforms LIN+LON, and fine-tuning outperforms both LIN+UP and LIN+LON. For letter sequence recognition in the next section, we adapt via fine-tuning using $20\%$ of the test signer's data.

\begin{figure*}[t]
\centering
\begin{tabular}{@{}c@{\hspace{0.02\linewidth}}c@{\hspace{0.01\linewidth}}c@{\hspace{0.01\linewidth}}c@{}}
No Adapt. & \caja{c}{c}{LIN+UP (Forced-Align.)} & \caja{c}{c}{LIN+UP (Ground Truth)} \\
\includegraphics[width=0.34\linewidth]{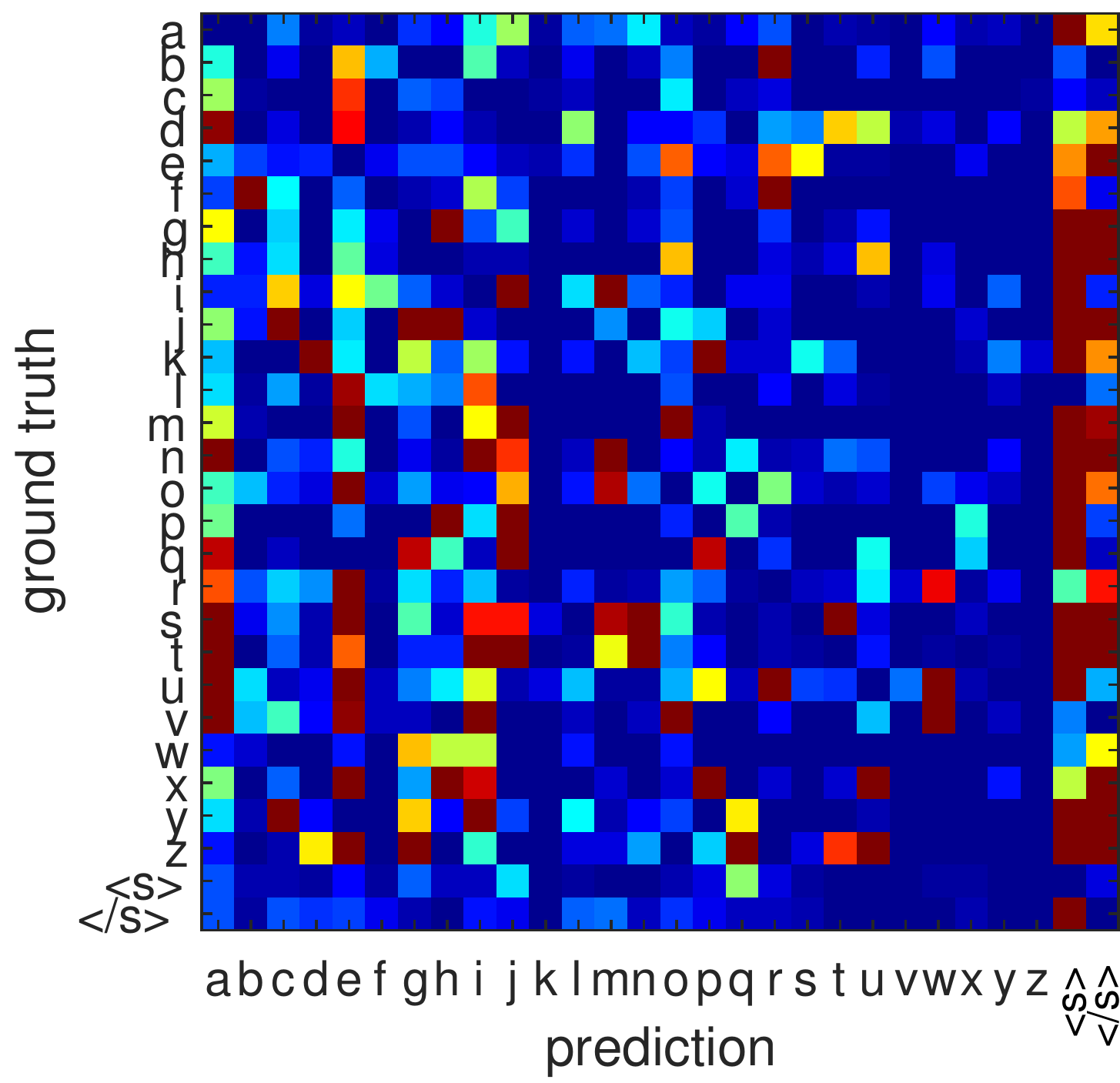} & 
\includegraphics[width=0.28\linewidth]{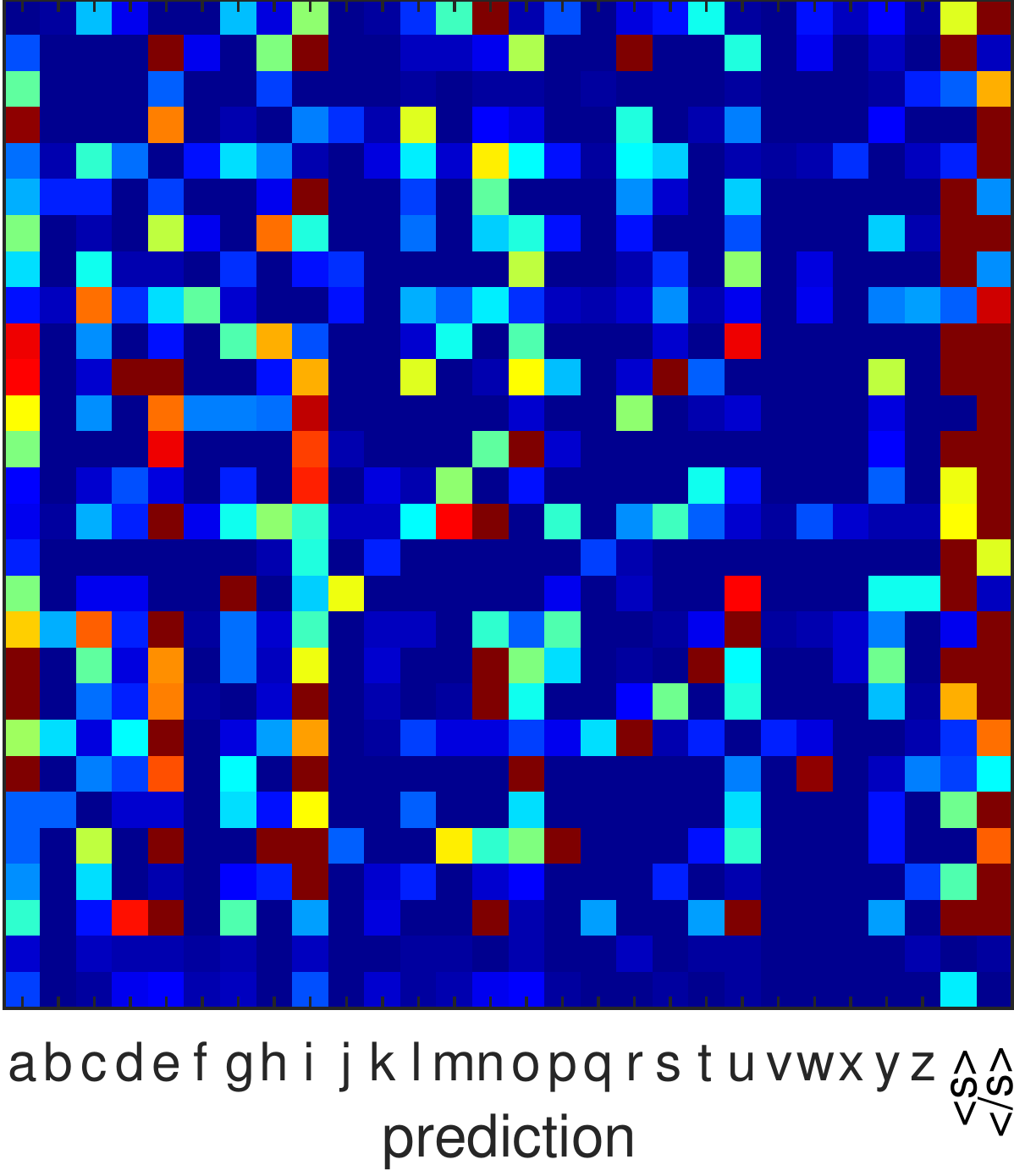} & 
\includegraphics[width=0.34\linewidth]{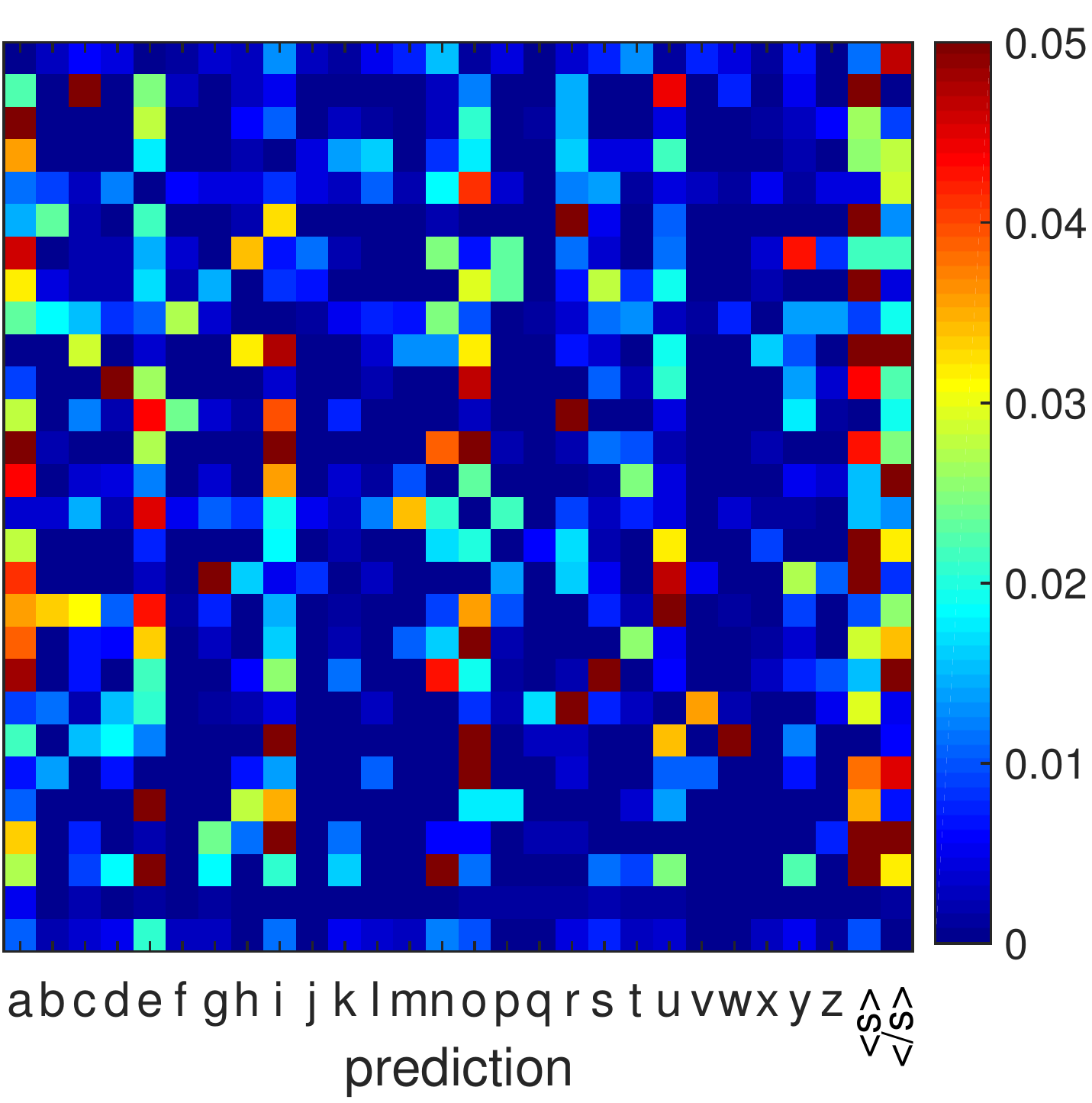} \\
\end{tabular}
\vspace{-.1in}
\caption{Confusion matrices of DNN classifiers for one test signer (Signer 1).  $20\%$ of the test signer's data (115 words) was used for adaptation, and a disjoint $70\%$ was used to compute confusion matrices.  
Each matrix cell is the empirical probability of the predicted class (column) given the ground-truth class (row).  The diagonal has been zeroed out for clarity.}
\label{fig:confmat}
\end{figure*}

Fig.~\ref{fig:confmat} further analyzes the DNNs via confusion matrices.  One of the main effects is the large number of incorrect predictions of the non-signing classes ($<$s$>$, $<$/s$>$).  We observe the same effect with the phonological feature classifiers.  This may be due to the previously mentioned fact that non-linguistic gestures are variable and easy to confuse with signing when given a new signer's image frames.  The confusion matrices show that, as the DNNs are adapted, this is the main type of error that is corrected.

\subsection{Connected letter recognition}

In connected letter recognition, we measure performance via the letter accuracy, analogously to the word or phone accuracy in speech recognition.  Table~\ref{t:LER} shows the letter accuracies obtained with the tandem, rescoring SCRF, and first-pass SCRF models with DNN adaptation via fine-tuning, using different types of adaptation data.  For all models, we do not retrain the models with the adapted DNNs, but tune several hyperparameters\footnote{See~\cite{kim2012,kim2013,tang2015} for details of the tuning parameters.} on 10\% of the test signer's data.  The tuned models are evaluated on an unseen 10\% of the test signer's remaining data; finally, we repeat this for eight choices of tuning and test sets, covering the 80\% of the test signer's data that we do not use for adaptation, and report the mean letter accuracy over the test sets.  


As shown in Table~\ref{t:LER}, without adaptation both tandem and SCRF models do poorly, achieving only roughly 40\% letter accuracies, with the rescoring SCRF slightly outperforming the others (recall that signer-dependent recognition achieves about 90\% letter accuracies~\cite{kim2013}).  With adaptation, however, performance jumps to up to 69.7\% letter accuracy with forced-alignment adaptation labels and up to 82.7\% accuracy with ground-truth adaptation labels.  All of the adapted models perform similarly, but interestingly, the first-pass SCRF is slightly worse than the others before adaptation and better (by 4.4\% absolute) after ground-truth adaptation.  One hypothesis is that the first-pass SCRF is more dependent on the DNN performance, while the tandem model uses the original image features and the rescoring SCRF uses the tandem model hypotheses and scores.  Once the DNNs are adapted, however, the first-pass SCRF outperforms the other models.


%% file: table.tex
\begin{table*}[ht!]
\centering
\resizebox{\linewidth}{!}{
\begin{tabular}{|l||c|c|c|c|c||c|c|c|c|c||c|c|c|c|c|}
\hline
        & \multicolumn{5}{c||}{Tandem HMM} & \multicolumn{5}{c||}{Rescoring SCRF} & \multicolumn{5}{c|}{1st-pass SCRF} \\ \hline
Signer  & 1 & 2 &  3 &  4 & {\bf Mean} &  1 &  2 &  3 &  4 & {\bf Mean} & 1 &  2 &  3 &  4 & {\bf Mean} \\ \hline\hline
No adapt. & 45.9&	45.3&	37.4&	42.5&	{\bf 42.8}&	47.4&	48.8&	38.9&	43.7&	{\bf 44.7}&	44.7&	46.7&	27.5&	38.6&	{\bf 39.4} \\ \hline
Forced align. & 69.8&    71.5&   60.4&   63.9&   {\bf 66.4}&     70.5&   74.0&   61.8&   65.5&   {\bf 68.0}&     75.6&   75.1&   63.5&   64.5&   {\bf 69.7} \\ \hline
Ground truth  & 78.0&    87.0&   68.4&   78.6&   {\bf 78.0}&     77.6&   86.5&   70.5&   78.6&   {\bf 78.3}&     84.8&   89.4&   75.1&   81.6&   {\bf 82.7} \\ \hline
\end{tabular}
}
\vspace{-.1in}
\caption{Letter accuracies (\%) on four test signers.}
\label{t:LER}
\end{table*}



%% file: conclusions.tex
\section{CONCLUSION}
\label{sec:conclusions}

In this study of signer-independent and adapted ASL fingerspelling recognition, we have seen that fingerspelling has great variability in speed, hand appearance, and appearance of non-signing gestures.  We have improved performance on new signers via adaptation of DNNs in tandem and SCRF recognizers.  Several DNN adaptation approaches are successful, with the largest improvements coming from simple fine-tuning on adaptation data.  This approach improves letter accuracies from around 40\% (unadapted) to up to 69.7\% with weak word-level supervision and up to 82.7\% with ground-truth frame labels for the adaptation data.  While the models perform similarly, the best adapted model is a first-pass SCRF.  The main DNN improvements come from resolving confusions between actual letters and the non-signing (``silence'') class.  Future work will continue to improve the models and adaptation approaches, as well as address other types of variability that are needed to port the models to video data ``in the wild''.

%% file: main.bbl
\begin{thebibliography}{10}

\bibitem{dreuw}
P.~Dreuw, D.~Rybach, T.~Deselaers, M.~Zahedi, and H.~Ney,
\newblock ``Speech recognition techniques for a sign language recognition
  system,''
\newblock in {\em Proc. Interspeech}, 2007.

\bibitem{zaki}
M.~M. Zaki and S.~I. Shaheen,
\newblock ``Sign language recognition using a combination of new vision based
  features,''
\newblock {\em Pattern Recognition Letters}, pp. 3397--3415, 2010.

\bibitem{bowden}
R.~Bowden, D.~Windridge, T.~Kadir, A.~Zisserman, and M.~Brady,
\newblock ``A linguistic feature vector for the visual interpretation of sign
  language,''
\newblock in {\em Proc. ECCV}, 2004.

\bibitem{liw}
S.~Liwicki and M.~Everingham,
\newblock ``Automatic recognition of fingerspelled words in {British Sign
  Language},''
\newblock in {\em Proc. 2nd IEEE Workshop on CVPR for Human Communicative
  Behavior Analysis}, 2009.

\bibitem{theo}
S.~Theodorakis, V.~Pitsikalis, and P.~Maragos,
\newblock ``Model-level data-driven sub-units for signs in videos of continuous
  sign language,''
\newblock in {\em Proc.~ICASSP}, 2010.

\bibitem{vog}
C.~Vogler and D.~Metaxas,
\newblock ``Toward scalability in {ASL} recognition: Breaking down signs into
  phonemes,''
\newblock in {\em Proc. Gesture Workshop}, 1999.

\bibitem{kim2012}
T.~Kim, K.~Livescu, and G.~Shakhnarovich,
\newblock ``American {S}ign {L}anguage fingerspelling recognition with
  phonological feature-based tandem models,''
\newblock in {\em Proc. SLT}, 2012.

\bibitem{kim2013}
T.~Kim, G.~Shakhnarovich, and K.~Livescu,
\newblock ``Fingerspelling recognition with semi-{M}arkov conditional random
  fields,''
\newblock in {\em Proc. ICCV}, 2013.

\bibitem{forster2013improving}
J.~Forster, O.~Koller, C.~Oberd{\"o}rfer, Y.~Gweth, and H.~Ney,
\newblock ``Improving continuous sign language recognition: {S}peech
  recognition techniques and system design,''
\newblock in {\em Proc. SLPAT}, 2013.

\bibitem{Padden}
C.~Padden and D.~C. Gunsauls,
\newblock ``How the alphabet came to be used in a sign language,''
\newblock {\em Sign Language Studies}, 2004.

\bibitem{bren}
D.~Brentari,
\newblock {\em A Prosodic Model of Sign Language Phonology},
\newblock MIT Press, 1998.

\bibitem{john}
R.~E. Johnson and S.~K. Liddell,
\newblock ``Toward a phonetic representation of signs: sequentiality and
  contrast,''
\newblock {\em Sign Language Studies}, vol. 11, no. 2, pp. 241--274, 2010.

\bibitem{athitsos2004boostmap}
V.~Athitsos, J.~Alon, S.~Sclaroff, and G.~Kollios,
\newblock ``Boost{M}ap: A method for efficient approximate similarity
  rankings,''
\newblock in {\em Proc. CVPR}, 2004.

\bibitem{Ricco-Tomasi-09}
S.~Ricco and C.~Tomasi,
\newblock ``Fingerspelling recognition through classification of
  letter-to-letter transitions,''
\newblock in {\em Proc. ACCV}, 2009.

\bibitem{Tsechpenakis-et-al-06-coupling}
G.~Tsechpenakis, D.~Metaxas, and C.~Neidle,
\newblock ``Learning-based dynamic coupling of discrete and continuous
  trackers,''
\newblock {\em Computer Vision and Image Understanding}, vol. 104, no. 2--3,
  2006.

\bibitem{Bowden-Sarhadi-02}
R.~Bowden and M.~Sarhadi,
\newblock ``A non-linear model of shape and motion for tracking finger spelt
  {A}merican sign language,''
\newblock {\em Image and Vision Computing}, vol. 20, no. 9--10, 2002.

\bibitem{Goh-Holden-06}
P.~Goh and E.~Holden,
\newblock ``Dynamic fingerspelling recognition using geometric and motion
  features,''
\newblock in {\em Proc. ICIP}, 2006.

\bibitem{Roussos:JMLR.13}
A.~Roussos, S.~Theodorakis, V.~Pitsikalis, and P.~Maragos,
\newblock ``Dynamic affine-invariant shape-appearance handshape features and
  classification in sign language videos,''
\newblock {\em Journal of Machine Learning Research}, vol. 14, 2013.

\bibitem{PugBow2011}
N.~Pugeault and R.~Bowden,
\newblock ``Spelling it out: Real-time {ASL} fingerspelling recognition,''
\newblock in {\em Proc. ICCV}, 2011.

\bibitem{ellis}
D.~P.~W. Ellis, R.~Singh, and S.~Sivadas,
\newblock ``Tandem acoustic modeling in large-vocabulary recognition,''
\newblock in {\em Proc. ICASSP}, 2001.

\bibitem{sarawagisemi}
S.~Sarawagi and W.~W. Cohen,
\newblock ``Semi-{M}arkov conditional random fields for information
  extraction,''
\newblock in {\em NIPS}, 2004.

\bibitem{zweig}
G.~Zweig and P.~Nguyen,
\newblock ``A segmental {CRF} approach to large vocabulary continuous speech
  recognition,''
\newblock in {\em Proc.~ASRU}, 2009.

\bibitem{tang2015}
H.~Tang, W.~Wang, K.~Gimpel, and K.~Livescu,
\newblock ``Discriminative segmental cascades for feature-rich phone
  recognition,''
\newblock in {\em Proc. ASRU}, 2015.

\bibitem{liao2013speaker}
H.~Liao,
\newblock ``Speaker adaptation of context dependent deep neural networks,''
\newblock in {\em Proc. ICASSP}, 2013.

\bibitem{abdel2013fast}
O.~Abdel-Hamid and H.~Jiang,
\newblock ``Fast speaker adaptation of hybrid {NN/HMM} model for speech
  recognition based on discriminative learning of speaker code,''
\newblock in {\em Proc. ICASSP}, 2013.

\bibitem{swietojanski2014learning}
P.~Swietojanski and S.~Renals,
\newblock ``Learning hidden unit contributions for unsupervised speaker
  adaptation of neural network acoustic models,''
\newblock in {\em Proc. SLT}, 2014.

\bibitem{doddipatla2014speaker}
R.~Doddipatla, M.~Hasan, and T.~Hain,
\newblock ``Speaker dependent bottleneck layer training for speaker adaptation
  in automatic speech recognition,''
\newblock in {\em Proc. Interspeech}, 2014.

\bibitem{Neto_95a}
J.~Neto, L.~Almeida, M.~Hochberg, C.~Martins, L.~Nunes, S.~Renals, and
  T.~Robinson,
\newblock ``Speaker-adaptation for hybrid {HMM-ANN} continuous speech
  recognition system,''
\newblock in {\em Proc. Eurospeech}, 1995.

\bibitem{Yao_12a}
K.~Yao, D.~Yu, F.~Seide, H.~Su, L.~Deng, and Y.~Gong,
\newblock ``Adaptation of context-dependent deep neural networks for automatic
  speech recognition,''
\newblock in {\em Proc. SLT}, 2012.

\bibitem{li2010}
B.~Li and K.~C. Sim,
\newblock ``Comparison of discriminative input and output transformations for
  speaker adaptation in the hybrid {NN/HMM} systems,''
\newblock in {\em Proc. Interspeech}, 2010.

\bibitem{hog}
N.~Dalal and B.~Triggs,
\newblock ``Histograms of oriented gradients for human detection,''
\newblock in {\em Proc. CVPR}, 2005.

\bibitem{Zeiler_13a}
M.~D. Zeiler, M.~Ranzato, R.~Monga, M.~Mao, K.~Yang, Q.~V. Le, P.~Nguyen,
  A.~Senior, V.~Vanhoucke, J.~Dean, and G.~E. Hinton,
\newblock ``On rectified linear units for speech processing,''
\newblock in {\em Proc. ICASSP}, 2013.

\bibitem{Srivas_14a}
N.~Srivastava, G.~E. Hinton, A.~Krizhevsky, I.~Sutskever, and R.~R.
  Salakhutdinov,
\newblock ``Dropout: {A} simple way to prevent neural networks from
  overfitting,''
\newblock {\em Journal of Machine Learning Research}, vol. 15, pp. 1929--1958,
  2014.

\end{thebibliography}
